\documentclass[letterpaper]{article} %
\usepackage{aaai2026} 
\usepackage{times} 
\usepackage{helvet} 
\usepackage{courier} 
\usepackage[hyphens]{url} 
\usepackage{graphicx}
\usepackage{natbib} 
\usepackage{caption}
\usepackage{booktabs}  
\usepackage{multirow}  
\usepackage{pifont}    
\usepackage[table]{xcolor} 

\usepackage{amsmath}
\usepackage[framemethod=default]{mdframed}
\definecolor{boxback}{rgb}{0.97,0.97,1.0} 
\newmdenv[
  backgroundcolor=boxback,
  linecolor=black,
  outerlinewidth=0.5pt,
  roundcorner=5pt,
  skipabove=\baselineskip,
  skipbelow=\baselineskip
]{boxedformula}

\usepackage{algorithm}
\usepackage{algorithmic}

\usepackage{newfloat}
\usepackage{listings}
\DeclareCaptionStyle{ruled}{labelfont=normalfont,labelsep=colon,strut=off} %
\floatstyle{ruled}
\newfloat{listing}{tb}{lst}{}
\floatname{listing}{Listing}
\title{Semantic Document Derendering: SVG Reconstruction via Vision-Language Modeling}
\author{
    Adam Hazimeh\textsuperscript{\rm 1}, 
    Ke Wang\textsuperscript{\rm 1}, 
    Mark Collier\textsuperscript{\rm 2}\equalcontrib, 
    Gilles Baechler\textsuperscript{\rm 2}\equalcontrib \\
    Efi Kokiopoulou\textsuperscript{\rm 2}\equalcontrib, 
    Pascal Frossard\textsuperscript{\rm 1}
}

\affiliations{

    \textsuperscript{\rm 1}EPFL,
    \textsuperscript{\rm 2}Google DeepMind \\
    adam.hazimeh@epfl.ch
}

\newcommand{\methodname}{\texttt{SliDer}}
\newcommand{\datasetname}{\texttt{Slide2SVG}}
\usepackage{booktabs}
\usepackage{multirow}
\usepackage{xcolor}
\usepackage{graphicx}
\usepackage{subcaption}
\usepackage{pifont}
\usepackage{adjustbox}

\begin{document}

\maketitle

\begin{figure*}[t]
    \centering
    \includegraphics[width=1\linewidth]{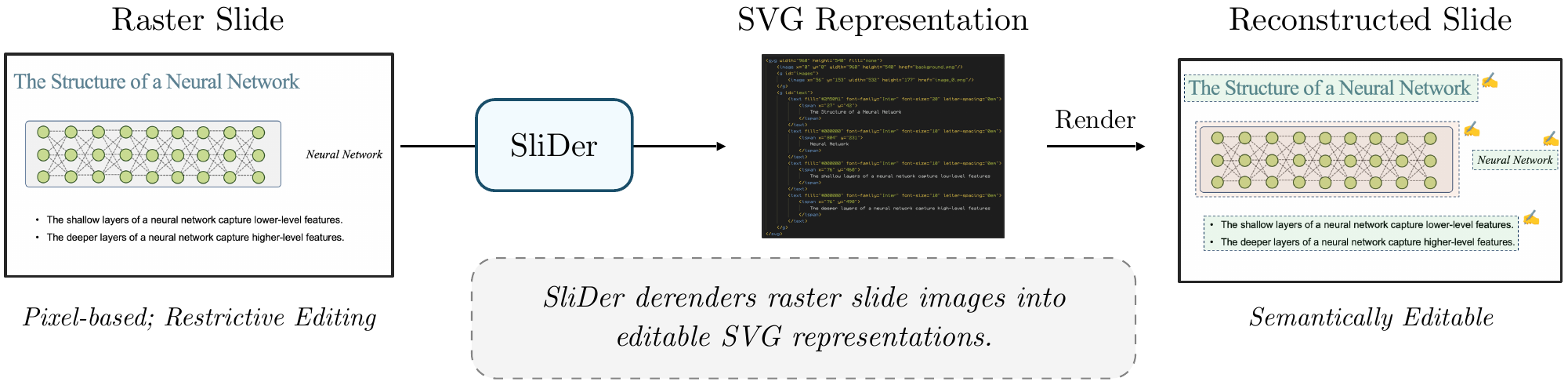}
    \caption{\methodname{} derenders raster slides into editable SVG-based format, allowing flexible editing on the slide such as adjusting figures, modifying text, etc.}
    \label{fig:frame_intro}
\end{figure*}


\begin{abstract}
Multimedia documents such as slide presentations and posters are designed to be interactive and easy to modify. Yet, they are often distributed in a static raster format, which limits editing and customization. Restoring their editability requires converting these raster images back into structured vector formats. However, existing geometric raster vectorization methods, which rely on low-level primitives like curves and polygons, fall short at this task. Specifically, when applied to complex documents like slides, they fail to preserve the high-level structure, resulting in a flat collection of shapes where the semantic distinction between image and text elements is lost. To overcome this limitation, we address the problem of \textit{semantic document derendering} by introducing \methodname{}, a novel framework that uses Vision-Language Models (VLMs) to derender slide images as compact and editable Scalable Vector Graphic (SVG) representations. \methodname{} detects and extracts the attributes from individual image and text elements in a raster input and organizes them into a coherent SVG format. Crucially, the model iteratively refines its predictions during inference in a process analogous to human design, generating SVG code that more faithfully reconstructs the original raster upon rendering. Furthermore, we introduce \datasetname{}, a novel dataset comprising raster-SVG pairs of slide documents curated from real-world scientific presentations, to facilitate future research in this domain. Our results demonstrate that \methodname{} achieves a reconstruction LPIPS of 0.069, and is favored by human evaluators in $82.9\%$ of cases compared to the strongest zero-shot VLM baseline. \looseness=-1
\end{abstract}

\begin{links}
    \link{Code \& dataset}{www.github.com/adamhazimeh/SliDer}
\end{links}


\section{Introduction}
\label{sec:intro}

Digital multimedia documents are often available as raster images, a format that conceals their underlying structure and hinders editability. To edit such documents, we must apply \textit{document derendering}, a process that first recovers the original layout from the pixel-based representation, and then parses identified assets like images and text, to semantically reconstruct the document into an editable form. This enables quick, accessible editing without the need to re-design documents from scratch.

Among common structured representation methods, Scalable Vector Graphics (SVGs) offer a flexible structure for representing multimedia documents by encoding image and text assets as discrete and editable elements.  Its hierarchical design allows for precise manipulation of individual components, facilitating straightforward editing and reordering.

Despite the advantages of SVG, most existing approaches that derender raster images into SVG format rely on low-level geometric primitives, such as curves and polygons \cite{ma2022towards, rodriguez2023starvector, carlier2020deepsvg, reddy2021im2vec}, which work well for simple icons and logos but fall short when applied to complex multimedia documents. These methods often produce unstructured representations that fail to capture the semantic layout of documents like slides, underscoring the need for SVG reconstruction techniques for multimedia documents that move beyond primitive-based approximations.

Overcoming these limitations requires a model that can interpret intricate visual inputs and generate structured code, a combination of capabilities that constitutes a core strength of modern Vision-Language Models (VLMs). Recent advancements in VLMs have showcased robust performance in code generation \cite{jiang2024survey, zheng2023survey} and image-to-text tasks \cite{team2023gemini, achiam2023gpt, bai2023qwen, team2025gemma}, demonstrating powerful image understanding and object detection abilities that are well-suited for high-level SVG reconstruction.

Motivated by these successes, we present \textbf{\methodname{}} (\textbf{Sli}de \textbf{Der}enderer), a novel VLM-based framework that converts raster multimedia documents into structured, editable SVG representations. We focus on slide-based documents, as their rich composition of text, images, and complex layouts makes them both a popular communication tool in many domains and a challenging benchmark. As illustrated in Figure \ref{fig:frame_intro}, our method derenders a raster slide into an SVG representation that faithfully reconstructs the original raster slide upon rendering. A key feature of our approach is its ability to iteratively refine its own predictions at inference time, allowing it to correct initial errors and progressively improve reconstruction fidelity. Notably, the images and text contained in the raster slide are parsed into individual, editable assets, enabling independent modifications.

To develop our method and advance research in this domain, we also introduce \textbf{\datasetname{}}, a new dataset for slide derendering. Comprising approximately 38,000 samples collected from real-world scientific presentations, it spans a wide array of designs, content, and layouts, providing a robust foundation for future work in structured document reconstruction.\looseness=-1

Using \datasetname{}, we evaluate \methodname{} with quantitative metrics and human judgments, focusing on the visual fidelity of its reconstructions. In pairwise tests, human evaluators chose Gemini-based \methodname{} over the strongest zero‑shot VLM baseline, GPT-4o \cite{hurst2024gpt}, in $82.9\%$ of cases and over LIVE \cite{ma2022towards}, a leading raster vectorization method, in $91.8\%$. Perceptual metrics also support this preference: \methodname{} achieves an LPIPS\footnote{LPIPS is a learned perceptual similarity metric in which lower values indicate higher visual similarity.} of $0.069$ compared to $0.118$ and $0.169$ for GPT-4o and LIVE, respectively, significantly reducing the perceptual distance between the original raster and the reconstruction.

The primary contributions of our work are as follows\footnote{Google DeepMind contributed in an advisory capacity only. No experiments or research were carried out by Google DeepMind.}:

\begin{itemize}
    \item We formulate the task of \textit{semantic document derendering}, which involves extracting the overall layout of a multimedia document and parsing each individual asset into an editable format, eventually transforming the raster document into a structured, editable representation.
    \item We propose \textbf{\methodname{}}, a VLM-based framework that iteratively converts raster slides into structured SVG representations, faithfully reconstructing the original slides upon rendering.\looseness=-1
    \item We introduce \textbf{\datasetname{}}, a novel dataset containing raster slides and their compact SVG representations, to address the shortage of image-to-SVG datasets for multimedia documents.
    \item We demonstrate through comprehensive quantitative and human evaluations on \datasetname{}, that \methodname{} consistently surpasses strong zero‑shot VLM and raster vectorization baselines in reconstruction fidelity.
\end{itemize}


\section{Related Work}
\label{sec:related_work}

We briefly survey work on vision-language models, raster vectorization, and document datasets most relevant to our setting, and refer the reader to the Appendix for an extended overview.

\subsection{Vision-Language Models}
Large Vision-Language Models (VLMs) have shown strong performance in image-to-text generation and visual reasoning \cite{li2025benchmark, zhang2024vision, hu2022scaling, xie2022visual, hartsock2024vision, lee2024visual}, including visual document understanding \cite{li2024enhancing, luo2022bi}. Because they can both parse complex layouts and generate structured code \cite{jiang2024survey, zheng2023survey}, they are a natural fit for SVG-based document derendering.

\subsection{Raster Vectorization}
Classical methods vectorize rasters via segmentation and diffusion curves \cite{selinger2003potrace, xia2009patch, orzan2008diffusion, xie2014hierarchical}, while more recent deep models such as DeepSVG, SVG-VAE, LIVE, Im2Vec, VectorFusion, and StarVector \cite{carlier2020deepsvg, lopes2019learned, ma2022towards, reddy2021im2vec, jain2023vectorfusion, rodriguez2023starvector} learn to generate or refine vector primitives. However, they typically output flat sets of paths and curves rather than a structured hierarchy of editable document elements, limiting their suitability for our task.\looseness=-1

\subsection{Datasets for SVG Generation and Document Understanding}
\label{sec:rel_work_datasets}
Existing SVG generation datasets mostly target simple graphics such as icons or emojis \cite{cai2023leveraging, wu2023iconshop, reddy2021im2vec, rodriguez2023starvector, cao2023svgformer}, and thus lack the layout complexity of real documents. Conversely, document understanding benchmarks like DocLayNet, PubLayNet, SlideVQA, and DocSynth \cite{pfitzmann2022doclaynet, zhong2019publaynet, tanaka2023slidevqa, zhao2024doclayout} focus on layout analysis but do not provide full, editable SVG representations. Our \datasetname{} dataset is designed to bridge this gap.


\begin{figure*}[ht!]
    \centering
    \includegraphics[width=1\linewidth]{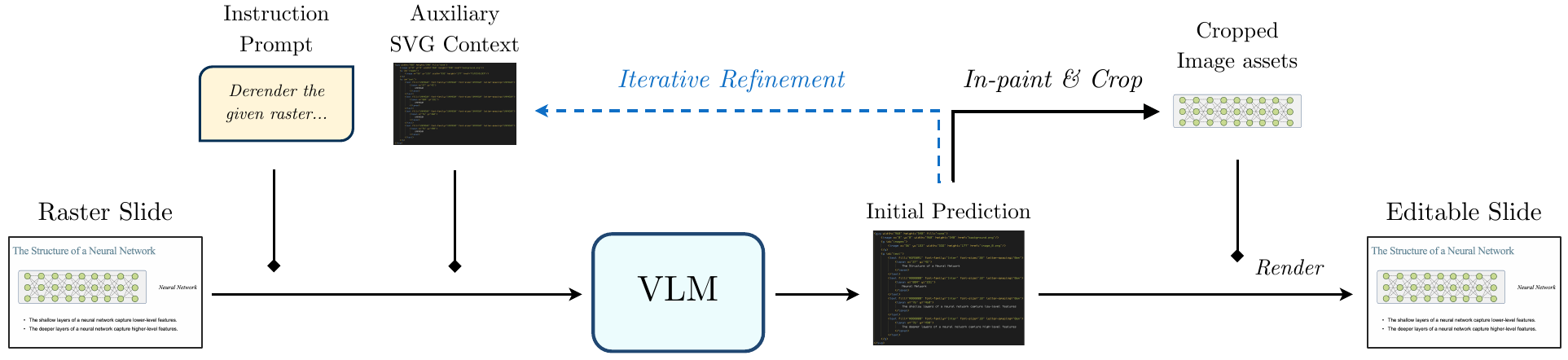}
    \caption{Overview of the \methodname{} inference pipeline. The VLM takes as input a raster slide, an instruction prompt, and an auxiliary SVG context, generating an editable SVG representation. The generated SVG can optionally be fed back to the VLM for iterative refinement. Given the final predicted SVG, the bounding box information is extracted to crop the image assets from the original raster into external PNG files. Finally, the slide is reconstructed by rendering the resulting SVG code.}
    \label{fig:pipeline}
\end{figure*}

\section{Background and Problem Formulation}
\label{sec:background}

\subsection{Representing Slides in SVG Format}

SVG offers a structured way to represent slide images by describing content as layered, discrete assets rather than as a dense array of pixels. It allows image and text elements to be encoded as individual objects with clearly defined attributes, providing layout-informed editability. For instance, image assets can be stored as external files referenced within the SVG, with attributes specifying their coordinates, width, and height. Similarly, text assets can be embedded directly into the SVG and include attributes such as the text content, position, font size, font family, color, and more.

This asset-based format provides fine-grained control over slide content: users can easily reposition figures, update text, and adjust layouts to meet evolving design requirements, making SVG an ideal candidate for semantic document derendering.\looseness=-1

\subsection{Slide Derendering Problem Formulation}

Slide derendering transforms a raster slide into a structured, editable SVG representation. This task poses three primary challenges:\looseness=-1

\begin{itemize} 
\item \textbf{Asset Identification:} Detecting and identifying individual elements, such as images and text, even in complex/overlapping layouts. \looseness=-1
\item \textbf{Attribute Inference:} Correctly determining each asset's attributes, including spatial coordinates and stylistic features.\looseness=-1
\item \textbf{SVG Code Generation:} Converting the inferred structure and attributes of identified assets into valid SVG code that faithfully replicates the original slide when rendered and supports further editing.
\end{itemize}

The ultimate goal of slide derendering is to produce SVG code that achieves two complementary objectives: faithfully replicating the visual design of the original slide, while representing the slide in a compact, easy-to-edit vector format.


\section{\textbf{\methodname{}}: A VLM-Based Framework for Slide Derendering}
\label{sec:framework}

To this end, we propose \textbf{\methodname{}}, a VLM-based approach that tackles the problem of slide derendering, converting raster slides into structured and editable SVG representations. It utilizes the visual understanding ability of VLMs to interpret the complex layout of the raster slide, identify the individual image and text assets, and extract their respective attributes. The extracted information is then organized by the VLM to generate a final SVG representation of the input raster slide. \looseness=-1

Importantly, derendering a raster slide is a complex task that requires accurately predicting the spatial attributes of individual assets, which is often difficult to accomplish in a single round of inference. To address this challenge, we integrate an \textit{iterative refinement} step into our framework. This enables the model to progressively improve its own predictions at inference time, correcting initial errors to achieve better reconstruction quality.

\subsection{Input Representation}
\label{sec:input_representation}

In our framework, the VLM takes three primary inputs during both training and inference: a raster slide, an instruction prompt, and an auxiliary SVG context.

\paragraph{Raster Slide}
The primary visual input is the raster slide that is to be derendered. Typically, a slide is composed of three types of assets:
\begin{itemize}
    \item \textit{Text boxes}, which represent text content as well as spatial ($x$, $y$, width, height) and stylistic attributes, including font size, font family, color, and letter spacing.
    \item \textit{Images}, e.g., figures, which contain RGB image content and spatial attributes. In the target SVG, image content is represented using an \verb|<href>| tag pointing to an external image file.\looseness=-1
    \item \textit{Background image}, which is assumed to stretch the canvas fully and is treated similarly to other image assets.
\end{itemize}

\paragraph{Derendering Instructions}
A text prompt provides high-level guidance, instructing the VLM to generate SVG code. The specific instruction used is: \looseness=-1
\begin{quote}
    ``De-render this raster image: \verb|<image>|. You may find the provided SVG template useful: \verb|<Auxiliary SVG Context>|"
\end{quote}

\paragraph{Auxiliary SVG Context}
To guide the VLM's prediction and enable iterative refinement, an SVG context is provided as auxiliary information. This context takes one of three forms, each serving a distinct purpose in training:
\begin{itemize}
    \item \textbf{Skeleton Template}: This is a bare-bones SVG structure containing no specific content, spatial, or stylistic attributes. Its purpose is to train the model to generate a complete SVG representation from scratch.
    \item \textbf{Partial Template}: This template contains only the spatial attributes (i.e., bounding box coordinates) for all text and image assets, with all stylistic attributes left empty. By providing the layout, this context focuses the model's task on learning to infer stylistic properties.
    \item \textbf{Initial Prediction}: This is a valid SVG representation generated by the same model architecture trained in a separate run, representing a first-pass prediction. It contains potentially imperfect spatial and stylistic attributes, and is crucial for training the model to perform iterative refinement by learning to correct its prior mistakes.
\end{itemize}

\subsection{Training Process}
\label{sec:training_process}

Based on the aforementioned inputs, a pre-trained, general-use VLM is fine-tuned to generate a complete SVG representation that faithfully reconstructs the provided raster image. The model produces the entire SVG code in a single generation step, by simultaneously predicting the spatial placement of image assets, which are hyper-referenced in the SVG code using \verb|<image>| tags, and the content and styling of text assets, leveraging its embedded OCR ability.

To ensure the model is robust and can handle different scenarios at inference time, we employ a data augmentation strategy that leverages the diverse input contexts described in Section \ref{sec:input_representation}. For raster slides in the training set, we create three distinct training variants by pairing them with different SVG contexts:
\begin{itemize}
    \item A \textit{skeleton template}, to learn generation from scratch.
    \item A \textit{partial template}, where bounding boxes are generated by an external YOLO-based object detector \cite{yolov8_ultralytics}.
    \item An \textit{initial prediction} (valid SVG), generated by a separately trained VLM. The starting context to obtain the initial prediction can either be a skeleton template, or a YOLO-guided partial template.
\end{itemize}
More details about the external models used are provided in the Appendix.

\subsection{Inference Process}
\label{sec:inference_process}

At inference time, \methodname{} follows a multi-step process to transform a raster slide into a fully editable SVG. The process begins with an initial generation pass, which can optionally be enhanced through iterative refinement, and concludes with a final post-processing step to extract all assets.

\subsubsection{Initial SVG Generation}
The process starts by feeding the VLM the input raster slide and an initial SVG context. This context is typically either a \textit{skeleton template} for generating the SVG from scratch, or a \textit{partial template} (i.e., with bounding boxes from an external object detector) to leverage prior spatial information.

\begin{table*}[ht]
\centering
\caption{Quantitative evaluation of different derendering methods. Zero-shot methods use YOLO-guided partial templates, with no iterative refinement, while results for \methodname{} are reported with iterative refinement. mIoU and OCR Accuracy are not available for LIVE since it only generates vector paths. Bold/\underline{underlined} values correspond to best/second-best per metric.}
\small
\setlength{\tabcolsep}{6pt} %
\begin{tabular}{@{}cccccccc@{}}
\toprule[1pt]
 &  & \multirow{2}{*}{mIoU (\%)$\,\uparrow$} & \multirow{2}{*}{\begin{tabular}[c]{@{}c@{}}OCR\\ Accuracy (\%)$\,\uparrow$\end{tabular}} & \multicolumn{3}{c}{Visual metrics} & \multirow{2}{*}{Elo $\,\uparrow$} \\
\cmidrule(l){5-7}
 &  &  &  & MSE$\,\downarrow$ & LPIPS$\,\downarrow$ & CLIP Sim.$\,\uparrow$ &  \\
\midrule[1pt]
\multirow{1}{*}{Raster Vectorization} 
 & LIVE                  & N/A      & N/A       & 18.19 & 0.169 & 0.7823 & 794   \\
\midrule
\multirow{3}{*}{Zero-shot VLMs}   
 & GPT-4o      & $88.42$  & $69.82$   & $14.48$ & $0.118$ & $0.883$ & 948 \\
 & Gemma      & $80.28$  & $65.57$   & $15.67$ & $0.150$ & $0.848$ & 880 \\
 & Gemini     & $83.78$  & $67.71$   & $15.05$ & $0.123$ & $0.879$ & 925 \\
\midrule
\multicolumn{1}{c}{\multirow{2}{*}{\textbf{\methodname{}}}}
 & Gemma                 & $\mathbf{89.36}$ & $\mathbf{93.53}$ & $\underline{13.38}$ & $\underline{0.075}$ & $\underline{0.950}$ & \underline{1207}      \\
\multicolumn{1}{c}{} 
 & Gemini                & $\underline{89.14}$  & $\underline{92.85}$   & $\mathbf{13.30}$ & $\mathbf{0.069}$ & $\mathbf{0.953}$ & $\mathbf{1245}$ \\
\bottomrule[1pt]
\end{tabular}
\label{tab:quantitative_results}
\end{table*}

\subsubsection{Iterative Refinement}
Complex slides with intricate layouts or subtle stylistic details can pose a challenge for any single-pass generation model. To enhance derendering quality in these cases, \methodname{} includes an optional iterative refinement capability. As shown in Figure \ref{fig:pipeline}, the SVG generated in the previous step can be fed back into the model as an \textit{initial} context, along with the original raster slide. This process allows the model to polish its initial prediction, addressing misalignments, stylistic inconsistencies, or layout errors. \looseness=-1

\subsubsection{Post-processing}
Once the final SVG representation is generated, we perform two post-processing steps to produce the final derendering.

\begin{itemize}
    \item \textbf{Background and Overlap Resolution:} This step resolves occlusions and isolates the background. If the predicted SVG bounding boxes for any assets overlap, we employ the TELEA \cite{telea2004image} inpainting algorithm to fill in the occluded regions. To extract the background, we mask out all foreground assets from the original raster and use the same inpainting technique to fill the resulting empty areas, producing a clean background image.

    \item \textbf{Image Asset Extraction:} After the inpainting step, all image assets defined in the final SVG are cropped from the original raster slide using their predicted bounding boxes. These cropped assets are then saved as external PNG files with filenames that correspond to those hyper-referenced in the generated SVG code (e.g., `\verb|image_1.png|').\looseness=-1
\end{itemize}

The final output of this entire process is a fully editable SVG file, accompanied by all associated image assets as standalone files.


\section{\datasetname{}: A New Dataset for Slide Derendering} 
\label{sec:slide2svg}

As established in Section \ref{sec:rel_work_datasets}, existing datasets for vector graphics and document understanding are ill-suited for the task of semantic document derendering. To fill this critical gap, we introduce \textbf{\datasetname{}}, a new real-world dataset designed to facilitate the transformation of rasterized slides into structured, editable SVG representations. Curated from publicly available conference presentations, the dataset captures diverse design styles, font choices, image placements, and layout configurations found in real-world slides, offering a challenging yet realistic platform for evaluating slide derendering pipelines. Each sample in \datasetname{} is composed of: \looseness=-1
\begin{itemize}
    \item The original raster slide in PNG format.
    \item The corresponding SVG representation of the raster slide, where text and image assets are encoded compactly to preserve editability.
    \item Individual image assets referenced in the SVG code, separately accessible as PNG files.
\end{itemize}

To construct this dataset, we assembled slides from academic conference presentations, particularly within the machine learning community, following a systematic data collection and processing pipeline:

\begin{enumerate}
    \item \textbf{PDF Collection} – We collect presentation slides in PDF format from the archives of several major machine learning conferences.
    \item \textbf{SVG Conversion} – The PDFs are then converted to Figma designs \cite{figma} and exported as raw SVG files. Figma is a web-based design tool primarily used for designing user interfaces and prototypes, but it also integrates community plugins, some of which can be used to convert PDFs to Figma designs. Note that this conversion is only used to build \datasetname{}. At inference time, we naturally assume that the slide is not available in a vector format (e.g., SVG, PDF) and must be derendered from a raster format.
    \item \textbf{Asset Grouping} – Text assets in the Figma-exported SVG slides are often arbitrarily grouped based on heuristics rather than semantic coherence. To ensure a structured and consistent grouping, we use the zero-shot DocLayout-YOLO model \cite{zhao2024doclayout} to reorganize text elements. Text assets identified as belonging to a single entity are merged into a unified text object, with their spatial attributes updated accordingly.
    \item \textbf{Outlier Filtering} – We filter out slides containing more than 8 image assets or 31 text assets (i.e., 95th percentile of asset counts), as they are excessively complex and not representative of common real-world slides.
    \item \textbf{Rasterization} – The obtained SVG representations are finally rendered into PNG format to obtain the corresponding raster slides.
\end{enumerate}

The final dataset is randomly divided into roughly 38,000 training samples and 225 test samples, each consisting of a raster image and its corresponding SVG representation.

By providing a new standardized dataset for raster-to-SVG conversion, \datasetname{} lays a foundation for future research in fields including document understanding and layout generation.\looseness=-1

\begin{figure*}[t]
    \centering
    \includegraphics[width=1\linewidth]{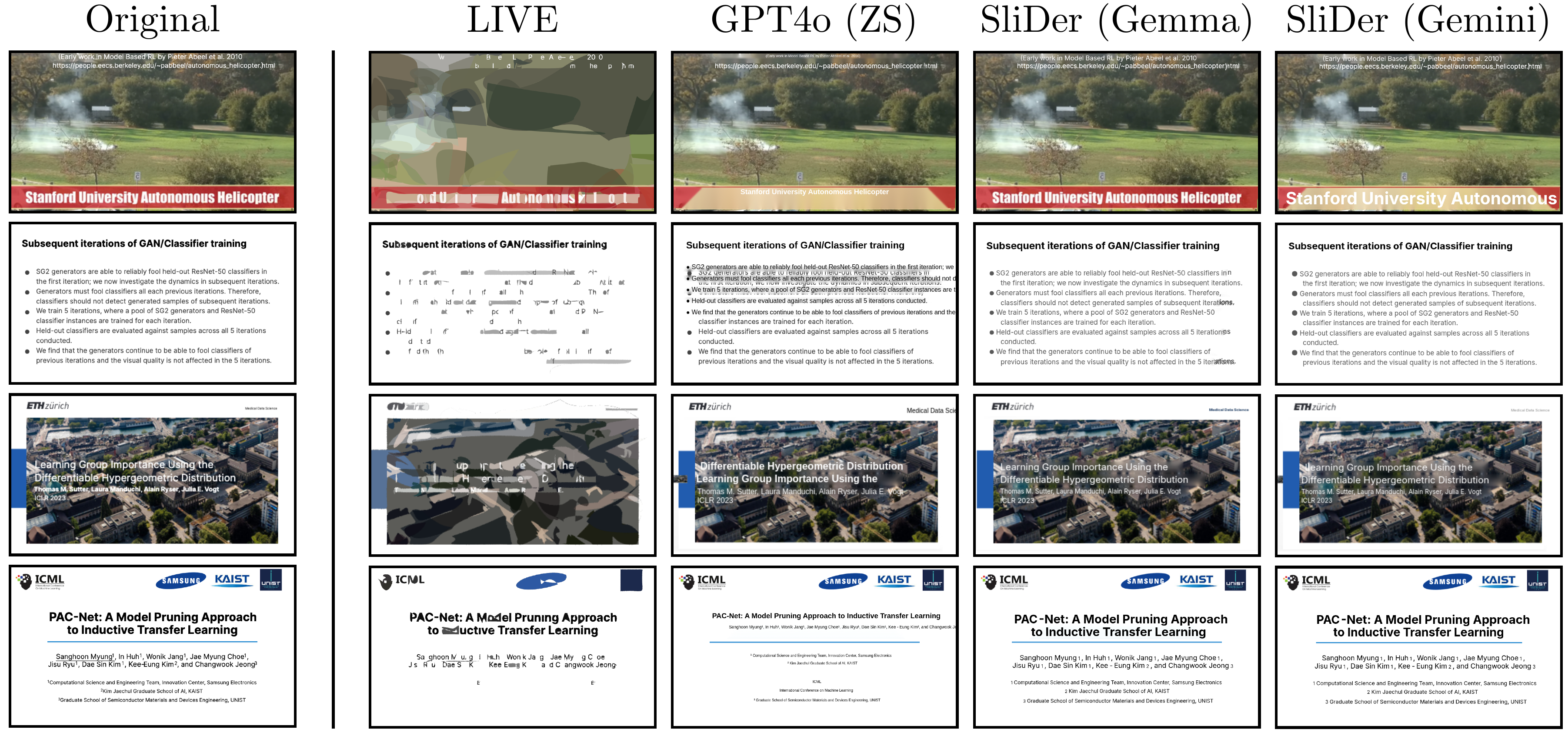}
    \caption{Examples of derendered slide images. Each row contains a separate sample, showing the original raster slide and the reconstructions from the derendered SVGs by different methods. For \methodname{}, we show the YOLO-guided versions with one step of iterative refinement. ``ZS" refers to zero-shot methods.}
    \label{fig:example_derender}
\end{figure*}


\section{Experiments}
\label{sec:experiments}

\subsection{Experimental Setup}

\paragraph{Models and Training Details}
We experiment with two large VLMs: Gemini-1.5-Flash \cite{team2024gemini} and the open-source Gemma 3 (12B) \cite{team2025gemma}. We train both models using the proposed \methodname{} framework on the training set of \datasetname{}. This framework is designed to train the model to handle diverse scenarios, such as generating SVGs from scratch (using skeleton templates), leveraging spatial priors (using partial templates), or refining previous outputs (using initial predictions). We demonstrate the results of \methodname{} configured with a partial template (bounding box priors) and iterative refinement in Table \ref{tab:quantitative_results}, while presenting other configurations in Table \ref{tab:ablation_results}.

The bounding box priors used in partial templates are obtained from a YOLOv8 model \cite{yolov8_ultralytics} trained for 50 epochs on the \datasetname{} training set, with the objective of detecting image and text objects in the raster slide. Initial predictions used for iterative refinement are generated by a VLM of the same family. For instance, the Gemini-based \methodname{} is refined using an initial prediction from a separately trained Gemini model. Further experimental details are presented in the Appendix.

\paragraph{Baselines}
We compare \methodname{} against two categories of baselines:
\begin{itemize}
    \item \textit{Zero-shot VLMs:} We evaluate the zero-shot performance of Gemini-1.5 Flash, Gemma 3 (12B), and GPT-4o \cite{hurst2024gpt}. To ensure a fair comparison, these models are also provided with the YOLO-guided partial template, but without iterative refinement
    \footnote{Our empirical results show that zero-shot VLMs do not benefit from iterative refinement.}.
    \looseness=-1
    \item \textit{Raster Vectorization:} We compare against LIVE \cite{ma2022towards}, a deep-learning-based raster vectorization method that generates low-level geometric primitives.
\end{itemize}

\subsection{Evaluation Metrics}
To systematically assess model performance, we evaluate the generated SVGs using metrics that measure varying aspects of derendering quality.
\begin{itemize}
    \item \textit{Bounding Box mIoU}: Measures the spatial alignment of localized assets by the mean Intersection over Union (mIoU) between predicted and ground-truth bounding boxes, averaged over both image and text elements. \looseness=-1
    \item \textit{Text OCR Accuracy}: Evaluates character-level similarity between predicted and ground-truth text, computed by concatenating all text into one string and measuring sequence-level accuracy.
    \item \textit{Mean Squared Error (MSE)}: Measures pixel-wise reconstruction error (smaller is better). MSE is computed on pixels in the range of [0, 255].
    \item \textit{CLIP Similarity} \cite{mayilvahanan2024does}: Computes the cosine similarity between image feature embeddings from a CLIP model ($\in$[-1, 1]; higher is better).
    \item \textit{Learned Perceptual Image Patch Similarity (LPIPS)} \cite{zhang2018lpips}: Uses deep features from VGG-16 \cite{vgg} to quantify human-aligned perceptual distance ($\in$[0, 1]; lower is better).
    \item \textit{Elo Score}: A rating system reflecting human preference. In our case, methods are compared pairwise, with human evaluators determining the superior output based on visual fidelity. These outcomes are then used to update each method's Elo rating, reflecting their relative performance (higher is better). To compute the Elo score, we collect rankings from 6 human evaluators and use an initial score of 1000 with a K-factor of 4, similar to the setup used in Chatbot Arena \cite{chatbotarena}.
\end{itemize}

\begin{figure*}[ht]
    \centering
    \includegraphics[width=0.9\linewidth]{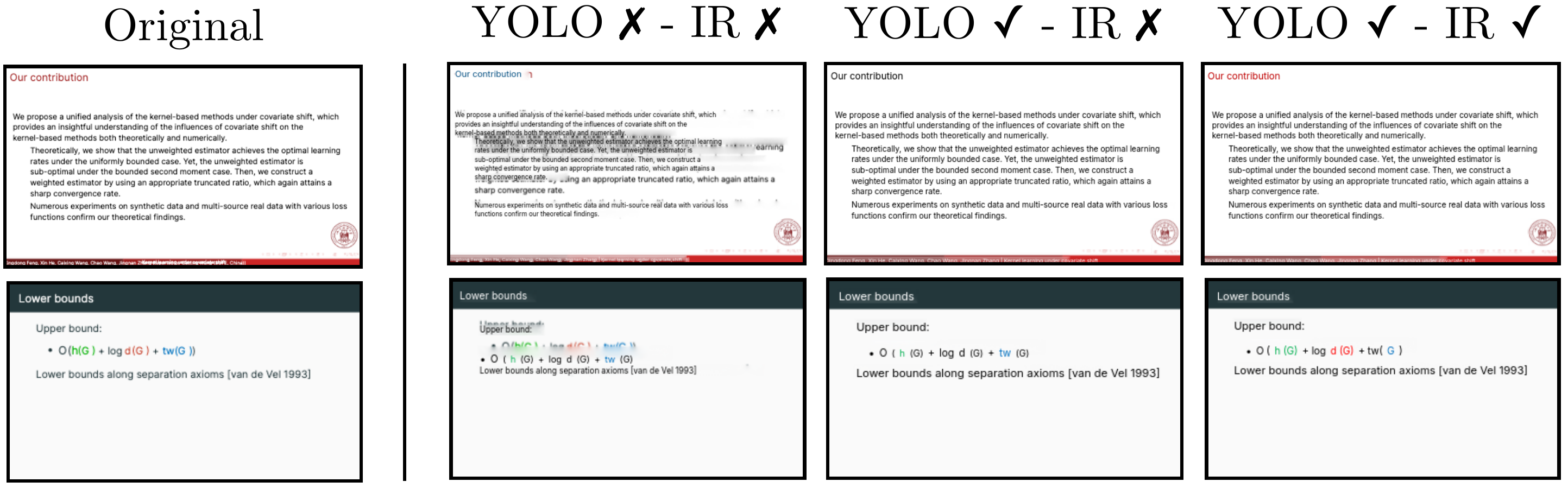}
    \caption{Qualitative examples for the ablations on the effect of bounding box information priors and iterative refinement during inference. We use the Gemini variant of \methodname{}. ``YOLO" indicates that the model uses bounding box priors. ``IR" indicates that the model performs one step of iterative refinement at inference time.}
    \label{fig:ablation_example}
\end{figure*}

\subsection{Main Results}
\label{sec:main_results}

Table \ref{tab:quantitative_results} presents the main quantitative evaluations. The results shows that our \methodname{} framework improves upon all baselines, with particularly notable gains in content preservation and perceptual quality as judged by both automated metrics and human evaluators.

The human evaluation results, measured by Elo scores, show a clear preference for our method. \methodname{} models score above 1200, significantly higher than 948 for the strongest zero-shot VLM, GPT-4o. This gap reflects that human evaluators favored our Gemini-based \methodname{} variant over GPT-4o in 82.9\% of cases, and over the geometric vectorization method (LIVE) in 91.8\% of cases. Details about computing the win-rate are presented in the Appendix.

This human preference aligns with the automated perceptual metrics. For instance, \methodname{}-Gemini achieves an LPIPS score of 0.069, a significant reduction 
compared to 0.118 for the best baseline. This is further supported by a higher CLIP Similarity score (0.953 vs. 0.883).

These perceptual improvements are driven by high fidelity in both content and layout. \methodname{} excels in text extraction, achieving an OCR accuracy above 93\%, a substantial increase from 69.82\% for the best baseline. This improvement arises because our fine-tuning process teaches the model to perform OCR effectively within the context of predefined bounding boxes. While the gain in layout accuracy (mIoU) is more modest, \methodname{} still outperforms the baselines. The lower mIoU of zero-shot VLMs is likely due to their tendency to make unnecessary layout edits and deviate from the provided templates. As expected, the geometric vectorization method, LIVE, is not competitive on this complex, multi-element derendering task.

\begin{table}[t!] %
\centering
\caption{Ablations for Gemini-based \methodname{}. We analyze the effect of adding YOLO-based bounding box priors (YOLO) and one step of iterative refinement (IR). mIoU and OCR accuracy are reported as percentages (\%). }
\small %
\setlength{\tabcolsep}{3.5pt} %
\renewcommand{\arraystretch}{1.2} %

\begin{tabular}{@{}c cc c c ccc@{}}
\toprule
& \multirow{2}{*}{YOLO} & \multirow{2}{*}{IR} & \multirow{2}{*}{mIoU↑} & \multirow{2}{*}{OCR↑} & \multicolumn{3}{c}{Visual metrics} \\
\cmidrule(l){6-8} 
& & & & & MSE↓ & LPIPS↓ & CLIP↑ \\
\midrule
\multirow{4}{*}{\shortstack{\textbf{\methodname{}}\\(Gemini)}} 
 & \textcolor{red!80!black}{\ding{55}} & \textcolor{red!80!black}{\ding{55}} & 81.90 & 89.78 & 14.47 & 0.090 & 0.929 \\
 & \textcolor{green!70!black}{\ding{51}} & \textcolor{red!80!black}{\ding{55}} & \textbf{89.71} & 92.42 & 13.46 & 0.072 & 0.951 \\
 & \textcolor{red!80!black}{\ding{55}} & \textcolor{green!70!black}{\ding{51}} & 81.70 & 90.57 & 14.28 & 0.088 & 0.930 \\
 & \textcolor{green!70!black}{\ding{51}} & \textcolor{green!70!black}{\ding{51}} & 89.14 & \textbf{92.85} & \textbf{13.30} & \textbf{0.069} & \textbf{0.953} \\ 
\bottomrule
\end{tabular}
\label{tab:ablation_results}
\end{table}

\subsection{Ablation Studies}
\label{sec:ablation_studies}

We conduct ablation studies on the Gemini-based \methodname{} model to isolate the effects of our framework's key components. The results are shown in Table \ref{tab:ablation_results}.

\paragraph{Effect of Prior Bounding Box Information}
Providing YOLO-based bounding box priors is shown to be effective for achieving a high-quality layout foundation. Comparing the model with priors (row 2) versus without (row 1), the mIoU jumps from 81.90\% to a stronger 89.71\%. This improved spatial localization has a cascading positive effect on other metrics, improving OCR accuracy from 89.78\% to 92.42\% and LPIPS from 0.090 to 0.072.  This validates that guiding the model with a reliable layout allows it to produce superior results, a practical strategy given that such priors can be readily obtained by pre-trained layout detection models like DocLayout-YOLO \cite{zhao2024doclayout}.
\paragraph{Effect of Iterative Refinement}
One step of iterative refinement provides a consistent boost in performance, acting as a final polishing step. When 
applied
to the model with YOLO priors (row 4 vs. row 2), we observe improvements across all visual metrics, with LPIPS dropping from 0.072 to 0.069. Even without priors (row 3 vs. row 1), refinement improves OCR accuracy and visual metrics. The best overall performance is achieved by combining both priors and refinement, validating our main model configuration choice. 

\subsection{Qualitative Analysis}
\label{sec:qualitative_analysis}

Figure \ref{fig:example_derender} provides a qualitative comparison of \methodname{} against LIVE and GPT-4o, where \methodname{}'s reconstructions closely replicate the original slides. In contrast, LIVE fails to render readable text, while the zero-shot GPT-4o output suffers from text misalignment and stylistic errors. Our method successfully captures fine-grained details, including logos and horizontal rules. These visual gains mirror the quantitative improvements reported in Table \ref{tab:quantitative_results}.\looseness=-1

Moreover, Figure \ref{fig:ablation_example} qualitatively demonstrates the impact of \methodname{}'s components. Without bounding box priors, text and image elements are visibly misaligned with their original locations. The introduction of YOLO priors corrects these spatial errors, creating a coherent layout. Finally, iterative refinement further reduces remaining stylistic inconsistencies and improves cropping quality. This visual progression confirms that each component plays a crucial role in achieving the final, high-fidelity reconstruction.\looseness=-1


\section{Conclusion \& Future Work} \label{sec:conclusion}
In this work, we presented \methodname{}, a VLM-based framework that transforms rasterized slides into structured, editable SVG representations. Our approach segments content elements and leverages an iterative refinement process to improve reconstruction fidelity. To facilitate this research, we also introduced \datasetname{}, a new dataset curated from real-world scientific presentations. Quantitative and human-preference evaluations confirm that \methodname{} produces editable outputs that are visually faithful and preferred over strong zero-shot baselines.

Opportunities for future work include improving performance on layouts with higher content density and expanding \methodname{} to support more complex multimedia documents, such as posters and infographics. Further investigation into the trade-offs between derendering quality and the computational cost of multi-step iterative refinement also presents a valuable research direction.


\section*{Acknowledgements}
We thank Alba Carballo Castro, Alessandro Favero, Amel Abdelraheem, Guillermo Ortiz-Jiménez, Imane Araf, Nikolaos Dimitriadis, Seth Nabarro, and Sevda Ögüt for helpful feedback and discussions.


\bibliography{aaai2026}

@String(CVPR= {IEEE Conf. Comput. Vis. Pattern Recog.})

@String(TOG= {ACM Trans. Graph.})

@String(ICLR = {Int. Conf. Learn. Represent.})

@String(AAAI = {AAAI})

@String(CVPR  = {CVPR})

@String(TOG   = {ACM TOG})

@String(ICLR  = {ICLR})

@misc{selinger2003potrace,
  title={Potrace: a polygon-based tracing algorithm},
  author={Selinger, Peter},
  url={https://potrace.sourceforge.net/potrace.pdf},
  year={2003}
}

@article{reddy2021im2vec,
  title={Im2Vec: Synthesizing Vector Graphics without Vector Supervision},
  author={Reddy, Pradyumna and Gharbi, Michael and Lukac, Michal and Mitra, Niloy J},
  journal={arXiv preprint arXiv:2102.02798},
  year={2021}
}

@article{rodriguez2023starvector,
  title={Starvector: Generating scalable vector graphics code from images},
  author={Rodriguez, Juan A and Agarwal, Shubham and Laradji, Issam H and Rodriguez, Pau and Vazquez, David and Pal, Christopher and Pedersoli, Marco},
  journal={arXiv preprint arXiv:2312.11556},
  year={2023}
}

@inproceedings{jain2023vectorfusion,
  title={Vectorfusion: Text-to-svg by abstracting pixel-based diffusion models},
  author={Jain, Ajay and Xie, Amber and Abbeel, Pieter},
  booktitle={IEEE/CVF Conference on Computer Vision and Pattern Recognition (CVPR)},
  year={2023}
}

@inproceedings{zhang2018LPIPS,
  title={The unreasonable effectiveness of deep features as a perceptual metric},
  author={Zhang, Richard and Isola, Phillip and Efros, Alexei A and Shechtman, Eli and Wang, Oliver},
  booktitle={IEEE Conference on Computer Vision and Pattern Recognition (CVPR)},
  year={2018}
}

@inproceedings{mayilvahanan2024does,
title={Does {CLIP}{\textquoteright}s generalization performance mainly stem from high train-test similarity?},
author={Prasanna Mayilvahanan and Thadd{\"a}us Wiedemer and Evgenia Rusak and Matthias Bethge and Wieland Brendel},
booktitle={International Conference on Learning Representations (ICLR)},
year={2024},
url={https://openreview.net/forum?id=tnBaiidobu}
}

@article{zhao2024doclayout,
  title={Doclayout-yolo: Enhancing document layout analysis through diverse synthetic data and global-to-local adaptive perception},
  author={Zhao, Zhiyuan and Kang, Hengrui and Wang, Bin and He, Conghui},
  journal={arXiv preprint arXiv:2410.12628},
  year={2024}
}

@article{achiam2023gpt,
  title={Gpt-4 technical report},
  author={Achiam, Josh and Adler, Steven and Agarwal, Sandhini and Ahmad, Lama and Akkaya, Ilge and Aleman, Florencia Leoni and Almeida, Diogo and Altenschmidt, Janko and Altman, Sam and Anadkat, Shyamal and others},
  journal={arXiv preprint arXiv:2303.08774},
  year={2023}
}

@article{team2023gemini,
  title={Gemini: a family of highly capable multimodal models},
  author={Team, Gemini and Anil, Rohan and Borgeaud, Sebastian and Alayrac, Jean-Baptiste and Yu, Jiahui and Soricut, Radu and Schalkwyk, Johan and Dai, Andrew M and Hauth, Anja and Millican, Katie and others},
  journal={arXiv preprint arXiv:2312.11805},
  year={2023}
}

@article{bai2023qwen,
  title={Qwen technical report},
  author={Bai, Jinze and Bai, Shuai and Chu, Yunfei and Cui, Zeyu and Dang, Kai and Deng, Xiaodong and Fan, Yang and Ge, Wenbin and Han, Yu and Huang, Fei and others},
  journal={arXiv preprint arXiv:2309.16609},
  year={2023}
}

@article{team2025gemma,
  title={Gemma 3 technical report},
  author={Team, Gemma and Kamath, Aishwarya and Ferret, Johan and Pathak, Shreya and Vieillard, Nino and Merhej, Ramona and Perrin, Sarah and Matejovicova, Tatiana and Ram{\'e}, Alexandre and Rivi{\`e}re, Morgane and others},
  journal={arXiv preprint arXiv:2503.19786},
  year={2025}
}

@inproceedings{ma2022towards,
  title={Towards layer-wise image vectorization},
  author={Ma, Xu and Zhou, Yuqian and Xu, Xingqian and Sun, Bin and Filev, Valerii and Orlov, Nikita and Fu, Yun and Shi, Humphrey},
  booktitle={IEEE/CVF Conference on Computer Vision and Pattern Recognition (CVPR)},
  year={2022}
}

@article{telea2004image,
  title={An image inpainting technique based on the fast marching method},
  author={Telea, Alexandru},
  journal={Journal of graphics tools},
  volume={9},
  number={1},
  pages={23--34},
  year={2004},
  publisher={Taylor \& Francis}
}

@article{team2024gemini,
  title={Gemini 1.5: Unlocking multimodal understanding across millions of tokens of context},
  author={Team, Gemini and Georgiev, Petko and Lei, Ving Ian and Burnell, Ryan and Bai, Libin and Gulati, Anmol and Tanzer, Garrett and Vincent, Damien and Pan, Zhufeng and Wang, Shibo and others},
  journal={arXiv preprint arXiv:2403.05530},
  year={2024}
}

@article{xia2009patch,
  title={Patch-based image vectorization with automatic curvilinear feature alignment},
  author={Xia, Tian and Liao, Binbin and Yu, Yizhou},
  journal={ACM Transactions on Graphics (TOG)},
  year={2009}
}

@article{orzan2008diffusion,
  title={Diffusion curves: a vector representation for smooth-shaded images},
  author={Orzan, Alexandrina and Bousseau, Adrien and Winnem{\"o}ller, Holger and Barla, Pascal and Thollot, Jo{\"e}lle and Salesin, David},
  journal={ACM Transactions on Graphics (TOG)},
  year={2008}
}

@article{xie2014hierarchical,
  title={Hierarchical diffusion curves for accurate automatic image vectorization},
  author={Xie, Guofu and Sun, Xin and Tong, Xin and Nowrouzezahrai, Derek},
  journal={ACM Transactions on Graphics (TOG)},
  year={2014}
}

@inproceedings{lopes2019learned,
  title={A learned representation for scalable vector graphics},
  author={Lopes, Raphael Gontijo and Ha, David and Eck, Douglas and Shlens, Jonathon},
  booktitle={IEEE/CVF International Conference on Computer Vision},
  year={2019}
}

@article{carlier2020deepsvg,
  title={Deepsvg: A hierarchical generative network for vector graphics animation},
  author={Carlier, Alexandre and Danelljan, Martin and Alahi, Alexandre and Timofte, Radu},
  journal={Advances in Neural Information Processing Systems (NeurIPS)},
  year={2020}
}

@article{li2025benchmark,
  title={Benchmark evaluations, applications, and challenges of large vision language models: A survey},
  author={Li, Zongxia and Wu, Xiyang and Du, Hongyang and Nghiem, Huy and Shi, Guangyao},
  journal={arXiv preprint arXiv:2501.02189},
  year={2025}
}

@article{zhang2024vision,
  title={Vision-language models for vision tasks: A survey},
  author={Zhang, Jingyi and Huang, Jiaxing and Jin, Sheng and Lu, Shijian},
  journal={IEEE Transactions on Pattern Analysis and Machine Intelligence},
  year={2024}
}

@article{hartsock2024vision,
  title={Vision-language models for medical report generation and visual question answering: A review},
  author={Hartsock, Iryna and Rasool, Ghulam},
  journal={Frontiers in Artificial Intelligence},
  volume={7},
  year={2024}
}

@article{lee2024visual,
  title={Visual question answering instruction: Unlocking multimodal large language model to domain-specific visual multitasks},
  author={Lee, Jusung and Cha, Sungguk and Lee, Younghyun and Yang, Cheoljong},
  journal={arXiv preprint arXiv:2402.08360},
  year={2024}
}

@inproceedings{hu2022scaling,
  title={Scaling up vision-language pre-training for image captioning},
  author={Hu, Xiaowei and Gan, Zhe and Wang, Jianfeng and Yang, Zhengyuan and Liu, Zicheng and Lu, Yumao and Wang, Lijuan},
  booktitle={IEEE/CVF Conference on Computer Vision and Pattern Recognition (CVPR)},
  year={2022}
}

@article{xie2022visual,
  title={Visual clues: Bridging vision and language foundations for image paragraph captioning},
  author={Xie, Yujia and Zhou, Luowei and Dai, Xiyang and Yuan, Lu and Bach, Nguyen and Liu, Ce and Zeng, Michael},
  journal={Advances in Neural Information Processing Systems (NeurIPS)},
  year={2022}
}

@inproceedings{li2024enhancing,
  title={Enhancing visual document understanding with contrastive learning in large visual-language models},
  author={Li, Xin and Wu, Yunfei and Jiang, Xinghua and Guo, Zhihao and Gong, Mingming and Cao, Haoyu and Liu, Yinsong and Jiang, Deqiang and Sun, Xing},
  booktitle={IEEE/CVF Conference on Computer Vision and Pattern Recognition (CVPR)},
  year={2024}
}

@article{luo2022bi,
  title={Bi-vldoc: Bidirectional vision-language modeling for visually-rich document understanding},
  author={Luo, Chuwei and Tang, Guozhi and Zheng, Qi and Yao, Cong and Jin, Lianwen and Li, Chenliang and Xue, Yang and Si, Luo},
  journal={arXiv preprint arXiv:2206.13155},
  year={2022}
}

@article{jiang2024survey,
  title={A survey on large language models for code generation},
  author={Jiang, Juyong and Wang, Fan and Shen, Jiasi and Kim, Sungju and Kim, Sunghun},
  journal={arXiv preprint arXiv:2406.00515},
  year={2024}
}

@article{zheng2023survey,
  title={A survey of large language models for code: Evolution, benchmarking, and future trends},
  author={Zheng, Zibin and Ning, Kaiwen and Wang, Yanlin and Zhang, Jingwen and Zheng, Dewu and Ye, Mingxi and Chen, Jiachi},
  journal={arXiv preprint arXiv:2311.10372},
  year={2023}
}

@article{cai2023leveraging,
  title={Leveraging large language models for scalable vector graphics-driven image understanding},
  author={Cai, Mu and Huang, Zeyi and Li, Yuheng and Ojha, Utkarsh and Wang, Haohan and Lee, Yong Jae},
  journal={arXiv preprint arXiv:2306.06094},
  year={2023}
}

@article{wu2023iconshop,
  title={Iconshop: Text-guided vector icon synthesis with autoregressive transformers},
  author={Wu, Ronghuan and Su, Wanchao and Ma, Kede and Liao, Jing},
  journal={ACM Transactions on Graphics (TOG)},
  year={2023}
}

@inproceedings{cao2023svgformer,
  title={Svgformer: Representation learning for continuous vector graphics using transformers},
  author={Cao, Defu and Wang, Zhaowen and Echevarria, Jose and Liu, Yan},
  booktitle={IEEE/CVF Conference on Computer Vision and Pattern Recognition (CVPR)},
  year={2023}
}

@inproceedings{pfitzmann2022doclaynet,
  title={Doclaynet: A large human-annotated dataset for document-layout segmentation},
  author={Pfitzmann, Birgit and Auer, Christoph and Dolfi, Michele and Nassar, Ahmed S and Staar, Peter},
  booktitle={ACM SIGKDD Conference on Knowledge Discovery and Data Mining},
  year={2022}
}

@inproceedings{zhong2019publaynet,
  title={Publaynet: largest dataset ever for document layout analysis},
  author={Zhong, Xu and Tang, Jianbin and Yepes, Antonio Jimeno},
  booktitle={International Conference on Document Analysis and Recognition (ICDAR)},
  year={2019}
}

@inproceedings{tanaka2023slidevqa,
  title={Slidevqa: A dataset for document visual question answering on multiple images},
  author={Tanaka, Ryota and Nishida, Kyosuke and Nishida, Kosuke and Hasegawa, Taku and Saito, Itsumi and Saito, Kuniko},
  booktitle={AAAI Conference on Artificial Intelligence},
  year={2023}
}

@misc{figma,
  author    = {{Figma, Inc.}},
  title     = {Figma - The Collaborative Interface Design Tool},
  year      = {2025},
  url       = {https://www.figma.com},
  note      = {Accessed: 2025-03-06}
}

@article{hurst2024gpt,
  title={Gpt-4o system card},
  author={Hurst, Aaron and Lerer, Adam and Goucher, Adam P and Perelman, Adam and Ramesh, Aditya and Clark, Aidan and Ostrow, AJ and Welihinda, Akila and Hayes, Alan and Radford, Alec and others},
  journal={arXiv preprint arXiv:2410.21276},
  year={2024}
}

@inproceedings{chatbotarena,
author = {Chiang, Wei-Lin and Zheng, Lianmin and Sheng, Ying and Angelopoulos, Anastasios N. and Li, Tianle and Li, Dacheng and Zhu, Banghua and Zhang, Hao and Jordan, Michael I. and Gonzalez, Joseph E. and Stoica, Ion},
title = {Chatbot arena: an open platform for evaluating LLMs by human preference},
year = {2024},
booktitle = {International Conference on Machine Learning}
}

@inproceedings{vgg,
  author       = {Karen Simonyan and
                  Andrew Zisserman},
  editor       = {Yoshua Bengio and
                  Yann LeCun},
  title        = {Very Deep Convolutional Networks for Large-Scale Image Recognition},
  booktitle    = {International Conference on Learning Representations (ICLR)},
  year         = {2015},
  url          = {http://arxiv.org/abs/1409.1556},
}

@software{yolov8_ultralytics,
  author = {Glenn Jocher and Ayush Chaurasia and Jing Qiu},
  title = {Ultralytics YOLOv8},
  version = {8.0.0},
  year = {2023},
  url = {https://github.com/ultralytics/ultralytics},
  orcid = {0000-0001-5950-6979, 0000-0002-7603-6750, 0000-0003-3783-7069},
  license = {AGPL-3.0}
}

\clearpage
\appendix
\onecolumn

\section*{Technical Appendix}
\addcontentsline{toc}{section}{Technical Appendix} %

\title{\textbf{Technical Appendix} \\ \Large{Semantic Document Derendering: SVG Reconstruction via Vision-Language Modeling}}
\date{}
\maketitle

\appendix

\section{Extended Related Work}
\subsection{Vision-Language Models}
Large Vision-Language Models (VLMs) have demonstrated remarkable capabilities in visual understanding and generating structured textual outputs \cite{li2025benchmark, zhang2024vision}. For example, VLMs have shown great performance in image-to-text generation tasks, such as image captioning \cite{hu2022scaling, xie2022visual} or visual question answering \cite{hartsock2024vision, lee2024visual}. Furthermore, VLMs have significantly advanced visual document understanding tasks, enabling the extraction of structured representations from scanned documents, forms, and academic papers \cite{li2024enhancing, luo2022bi}. Since VLMs possess both the ability to interpret complex layouts and to generate usable code \cite{jiang2024survey, zheng2023survey}, they are well-suited to address the challenges of asset parsing and SVG generation inherent to document derendering.

\subsection{Raster Vectorization}
Traditional raster vectorization methods used techniques like segmentation \cite{selinger2003potrace, xia2009patch} and diffusion curves \cite{orzan2008diffusion, xie2014hierarchical} to convert rasters to vector representations. More recently, deep learning-based methods like DeepSVG \cite{carlier2020deepsvg} and SVG-VAE \cite{lopes2019learned} employed hierarchical or latent variable models to learn vector primitives from datasets of icons and fonts. LIVE \cite{ma2022towards} and Im2Vec \cite{reddy2021im2vec} iteratively refine vector paths to match input rasters. More recent works take advantage of advances in generative models. For instance, VectorFusion \cite{jain2023vectorfusion} trains a text-conditioned diffusion model to vectorize rasters, while StarVector \cite{rodriguez2023starvector} aligns image encodings with a large language model to generate SVG code. Importantly, these methods produce a flat set of geometric primitives (e.g., paths, curves) instead of a hierarchical structure of editable semantic components, making them unsuitable for document derendering.\looseness=-1

\subsection{Datasets for SVG generation and Document Understanding}
Existing datasets for SVG generation primarily focus on vectorization of simple raster images such as icons or emojis \cite{cai2023leveraging, wu2023iconshop, reddy2021im2vec, rodriguez2023starvector, cao2023svgformer}. These datasets lack the structural complexity and layout diversity found in real-world documents, limiting their applicability in understanding and derendering scientific documents. Moreover, while several real-world datasets exist for document understanding, they are not designed for SVG-based document reconstruction and do not contain all necessary information for derendering. For example, while datasets such as DocLayNet \cite{pfitzmann2022doclaynet}, PubLayNet \cite{zhong2019publaynet}, SlideVQA \cite{tanaka2023slidevqa}, and DocSynth \cite{zhao2024doclayout} provide valuable resources for layout analysis and document parsing, they lack structured text representations and stylistic attributes like fonts and precise positional information required for SVG reconstruction.

\section{Framework Details}

\subsection{Model Architectures and Training Hyperparameters}
\paragraph{\methodname{} (Gemini-1.5-Flash)} The Gemini-1.5-Flash\footnote{Google DeepMind. `Gemini". \url{https://cloud.google.com/vertex-ai/generative-ai/docs/models/gemini/1-5-flash}} model was fine-tuned for one epoch on Google Cloud Platform's Vertex AI tuning service, which uses Parameter-Efficient Fine-tuning (PEFT) with a default adapter size of 8. Other specific training hyperparameters, such as learning rate and optimizer details, are managed by the platform and are not exposed to the user.

\paragraph{\methodname{} (Gemma 3)} The instruction-tuned Gemma 3 (12B)\footnote{Google DeepMind. `Gemma". \url{https://deepmind.google/models/gemma/gemma-3/}} model was fine-tuned locally for one epoch using LoRA. The key hyperparameters are detailed below:
\begin{itemize}
    \item \textbf{LoRA Rank:} 8
    \item \textbf{Maximum Sequence Length:} 8192 tokens
    \item \textbf{Batch Size:} 8 (1 per GPU with 8 steps of gradient accumulation)
    \item \textbf{Learning Rate:} 1e-4
    \item \textbf{Learning Rate Scheduler:} Cosine schedule with a warmup ratio of 0.1
    \item \textbf{Optimizer:} AdamW ($\beta_1=0.9, \beta_2=0.999$)
    \item \textbf{Hardware:} 8 NVIDIA H100 GPUs
    \item \textbf{Software:} LLaMA-Factory\footnote{Y. Zheng et al. LLaMA-Factory. \url{https://github.com/hiyouga/LLaMA-Factory}}
\end{itemize}

\subsection{Training Data Composition}
To ensure robust performance across different inference scenarios, we fine-tuned our models on an augmented dataset. Due to computational and cost constraints, all reported models were trained on a random 20,000-sample subset of the \datasetname{} training set. For each sample in this subset, we generated three distinct training variants by pairing it with different auxiliary SVG contexts, producing a total of 60,000 training samples:
\begin{itemize}
    \item \textbf{Skeleton Templates (20,000 instances):} Samples paired with empty SVG structures to teach generation from scratch.
    \item \textbf{Partial Templates (20,000 instances):} Samples paired with YOLO-guided bounding boxes to teach generation from a spatial prior.
    \item \textbf{Initial Predictions (20,000 instances):} To teach iterative refinement, we used half of the 20,000 sample subset (i.e., 10,000 samples) to generate two initial predictions: one starting from a skeleton context and the other from a partial context. This resulted in a total of 20,000 training instances for this task. The full 20,000 subset was not used so that we maintain an equal mixing ratio and avoid overpowering the other contexts.
\end{itemize}
The Gemma-based model was trained on the same data configuration for fairness.

\subsection{Bounding Box Priors}
The spatial priors for our partial templates were generated using a pre-trained YOLOv8n model\footnote{Ultralytics YOLOv8, \url{https://github.com/ultralytics/ultralytics}} that was then fine-tuned on our data. Key details include:
\begin{itemize}
    \item \textbf{Fine-tuning:} The model was fine-tuned for 50 epochs on the \datasetname{} training set.
    \item \textbf{Hyperparameters:} Default hyperparameters were used (learning rate of 0.01 and batch size of 16).
    \item \textbf{Output Classes:} The model was trained to detect two classes: `image' and `text'.
\end{itemize}

\subsection{Iterative Refinement and Post-Processing}
\paragraph{Generation of Initial Predictions} The ``initial predictions" used for refinement training were generated by separately trained VLM instances. For example, to create the refinement data for the final \methodname{}-Gemini model, we first fine-tuned two separate Gemini models: one trained solely on skeleton templates and another on partial templates. We then ran inference with these models on the training data to produce the corpus of initial predictions that the final model learns to correct.
\paragraph{Post-Processing} The background isolation and occlusion resolution step described in the main paper uses the TELEA inpainting algorithm\footnote{A. Telea. An image inpainting technique based on the fast marching method. \emph{Journal of Graphics Tools}, 9(1):23--34, 2004.} with its default parameters as implemented in OpenCV.

\section{\datasetname{} Details}

\subsection{Data Sources and Initial Conversion}
\paragraph{PDF Sources} The dataset was curated from presentation slides in PDF format, collected from the public archives of four major machine learning conferences: NeurIPS, ICML, ICLR, and CVPR.
\paragraph{PDF-to-SVG Conversion} The collected PDFs were converted into editable vector graphics using the \textbf{pdf.to.design} community plugin for Figma\footnote{Divriots. pdf.to.design Figma Plugin. \url{https://www.figma.com/community/plugin/1280917768965269588/pdf-to-design-by-divriots-import-any-pdf-to-figma}}. The resulting designs were then exported as raw SVG files for further processing.
\paragraph{Image Resolution} Final raster images (PNG) in the dataset have variable resolutions. During processing, they are resized to a maximum of 1024px on their longer side while preserving the aspect ratio.

\subsection{Dataset Statistics}
\label{sec:dataset_statistics}

To better characterize \datasetname{}, we report basic statistics over the
slides in the corpus.

\begin{figure}[t]
    \centering
    \includegraphics[width=\linewidth]{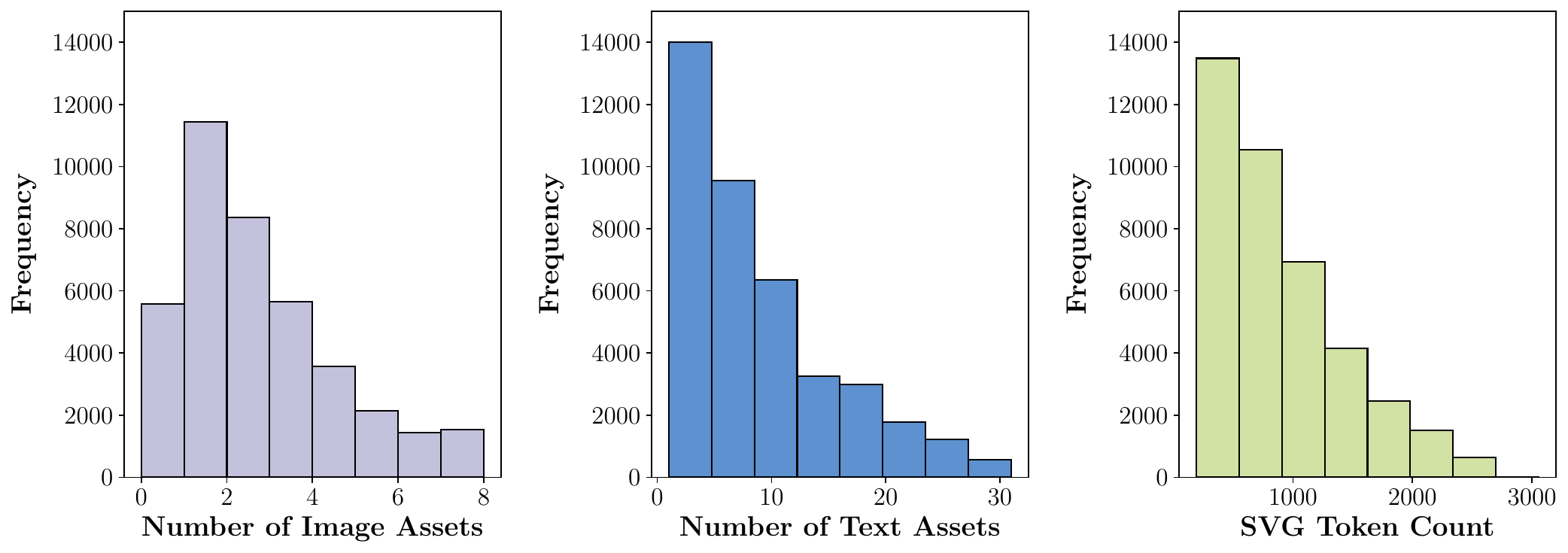}
    \caption{
    Distributions of (left) number of image assets per slide, (middle)
    number of text assets per slide, and (right) SVG token count in
    \datasetname{}. Most slides contain a small number of images and a
    moderate number of text elements, with a tail of more complex slides.}
    \label{fig:dataset_statistics}
\end{figure}

Figure~\ref{fig:dataset_statistics} summarizes three important aspects of the
data: (1) the number of image assets per slide, (2) the number of text assets per
slide, and (3) the total SVG token count. Most slides contain only a small
number of image assets (typically 1-3), reflecting the
common design pattern of combining a few key figures with text.\looseness=-1

The middle panel shows that the number of text assets per slide is more
variable, but still concentrated in a moderate range (e.g., titles, bullet
points, axis labels, short annotations, variable-style text, etc.), with a long tail of more
text-heavy slides. Finally, the right panel reports the distribution of
token counts (Gemini tokenizer) of the ground-truth SVGs, indicating that the majority of
slides fall within a relatively compact range of SVG length, with a gradual
tail towards more complex layouts. Overall, these statistics indicate that
\datasetname{} covers both simple and moderately complex slide designs,
while remaining within a regime that is practical for VLM-based SVG
generation. \looseness=-1

\subsection{Ground-Truth SVG Processing Pipeline}
The raw SVGs exported from Figma were processed through an automated cleaning and normalization pipeline to create high-quality ground-truth data. The key steps include:
\begin{enumerate}
    \item \textbf{Rasterization of Vector Shapes:} All non-image vector shapes (e.g., rectangles, circles) were automatically rendered into PNG images using a headless Firefox browser instance and then re-integrated into the SVG.
    \item \textbf{SVG Flattening and Cleanup:} The nested hierarchy of the original SVG was flattened, and all Figma-generated groupings were removed.
    \item \textbf{Asset Grouping:} Image assets were grouped based on proximity. Text assets were grouped using bounding boxes predicted by the pre-trained DocLayout-YOLO model, with any overlapping text boxes merged into a single coherent element.
    \item \textbf{Coordinate System Normalization:} The coordinate system was standardized to use relative positioning (percentages), which is more stable for model learning than absolute pixel values.
\end{enumerate}

\subsection{Auxiliary SVG Contexts}
Below are conceptual examples of the three types of auxiliary SVG contexts used to guide the VLM, for the same sample.
\begin{lstlisting}[language=XML, caption={A \textbf{skeleton template}, providing only the basic SVG structure for generation from scratch.}, label=lst:skeleton]
<svg xmlns="http://www.w3.org/2000/svg" xmlns:xlink="http://www.w3.org/1999/xlink" width="792" height="612" fill="white">
  <image x="0.0%
  <g id="images">
    <image x="UNKNOWN" y="UNKNOWN" width="UNKNOWN" height="UNKNOWN" href="image.png" />
    ...
  </g>
  <g id="text">
    <foreignObject x="UNKNOWN" y="UNKNOWN" width="UNKNOWN" height="UNKNOWN" overflow="visible">
      <div xmlns="http://www.w3.org/1999/xhtml" style="font-family: UNKNOWN; letter-spacing: UNKNOWN; font-weight: UNKNOWN; color: UNKNOWN; text-align: UNKNOWN;">
        <div>
          UNKNOWN
        </div>
      </div>
    </foreignObject>
    ...
  </g>
</svg>
\end{lstlisting}
\begin{lstlisting}[language=XML, caption={A \textbf{partial template}, providing pre-detected bounding boxes. The model must infer content and style.}, label=lst:partial]
<svg xmlns="http://www.w3.org/2000/svg" xmlns:xlink="http://www.w3.org/1999/xlink" width="720" height="405" fill="white">
  <image x="0.0%
  <g id="images">
    <image x="35.3%
  </g>
  <g id="text">
    <foreignObject x="1.8%
      <div xmlns="http://www.w3.org/1999/xhtml" style="font-family: UNKNOWN; font-size: UNKNOWN; letter-spacing: UNKNOWN; font-weight: UNKNOWN; color: UNKNOWN; text-align: UNKNOWN;">
        <div>UNKNOWN</div>
      </div>
    </foreignObject>
  </g>
</svg>
\end{lstlisting}
\begin{lstlisting}[language=XML, caption={An \textbf{initial prediction}, representing a plausible but imperfect SVG that the refinement model learns to correct.}, label=lst:initial]
<svg xmlns="http://www.w3.org/2000/svg" xmlns:xlink="http://www.w3.org/1999/xlink" width="720" height="405" fill="white">
  <image x="0.0%
  <g id="images">
    <image x="35.3%
  </g>
  <g id="text">
    <foreignObject x="1.8%
      <div xmlns="http://www.w3.org/1999/xhtml" style="font-family: Inter; font-size: 24px; letter-spacing: 0.0em; color: #000000; text-align: left;">
        <div>MOLERE Algorithm</div>
      </div>
    </foreignObject>
  </g>
</svg>
\end{lstlisting}

\section{Evaluation Details}

\subsection{Baseline Implementation}
\paragraph{LIVE} We used the official implementation of LIVE, available from its public GitHub repository\footnote{Picsart AI Research. LIVE: Layer-wise Image Vectorization. \emph{GitHub Repository}. \url{https://github.com/Picsart-AI-Research/LIVE-Layerwise-Image-Vectorization}}. For our experiments, we configured it with an exponential path schedule (base 2), a maximum of 128 total paths, and 32 paths per optimization iteration. This is the highest-compute configuration available in the LIVE repository.

\paragraph{Zero-shot VLMs} The zero-shot GPT-4o baseline\footnote{OpenAI.''GPT-4o". \url{https://platform.openai.com/docs/models/gpt-4o}} was accessed via the OpenAI API. Similarly, the zero-shot Gemini baseline utilized the Vertex AI inference API. For the zero-shot Gemma baseline, we used a locally hosted instruction-tuned Gemma 3 (12B) model without further fine-tuning. For all three models, default generation hyperparameters were used.

\subsection{Quantitative Metric Implementation}
Our evaluation metrics are implemented as follows:

\paragraph{Bounding Box mIoU}
This metric measures the spatial alignment of predicted assets (images and text) against the ground truth. To ensure a balanced assessment, we compute a symmetric Intersection over Union (IoU) for each asset class, which is designed to be fair to different kinds of errors.\\\\
This symmetric IoU is calculated by averaging two directional coverage scores:

\begin{enumerate}
    \item \textbf{Ground-Truth Coverage:} To see how well the predictions cover the ground truth, we measure, for each ground-truth box, what fraction of its area is overlapped by any of the predicted boxes. These fractions are then averaged across all ground-truth boxes. This score is high when the model successfully finds and places all ground-truth assets.
    
    \item \textbf{Prediction Coverage:} To measure how accurate the predictions are, we measure, for each predicted box, what fraction of its area overlaps with any of the ground-truth boxes. This score is then averaged across all predicted boxes. This score is high when the model does not generate extra or misplaced assets.
\end{enumerate}

\noindent The symmetric IoU for a given asset class (e.g., $\text{IoU}_{\text{text}}$) is the average of these two coverage scores. The final mIoU score is the mean of the symmetric IoUs for both text and image assets:
\begin{boxedformula}
    \begin{equation}
        \text{mIoU} = \frac{1}{2} (\text{IoU}_{\text{text}} + \text{IoU}_{\text{image}})
    \end{equation}
\end{boxedformula}

\paragraph{OCR Accuracy}
This metric evaluates character-level text fidelity and penalizes incorrect text ordering. First, all text content from the ground-truth SVG is extracted and concatenated into a single string, $S_{gt}$. The same is done for the predicted SVG to create $S_{pred}$. The order of text elements is preserved as it appears in the SVG's Document Object Model (DOM).\\

\noindent The accuracy is then defined as the normalized Levenshtein similarity, which measures the character-level edit distance between the two strings and normalizes it by the length of the longer string:
\begin{boxedformula}
    \begin{equation}
        \text{Accuracy} = 1 - \frac{\text{Levenshtein}(S_{gt}, S_{pred})}{\max(|S_{gt}|, |S_{pred}|)}
    \end{equation}
\end{boxedformula}
where $\text{Levenshtein}(\cdot)$ is the function to compute minimal edit distance. This score is bounded between 0 and 1, with 1 indicating a perfect match in both content and order.

\subsection{Human Evaluation Protocol}
\paragraph{Ranking and Elo Calculation} Human evaluations were performed using a custom interface where participants ranked the outputs of six methods. These full rankings were then decomposed into all constituent pairwise comparisons, and the Elo ratings were updated based on these pairwise outcomes. 

\paragraph{Win rates} The win rate for a method $A$ against another method $B$, also known as its expected score, is the probability of $A$ winning over $B$. To approximate the win rates reported in the main paper, the final Elo scores of the two methods can be plugged into the following formula\footnote{Elo, A. E. (1978). \emph{The Rating of Chessplayers, Past and Present}. Arco Publishing.}:
\begin{boxedformula}
\begin{equation}
\text{Win Rate}_{\text{A vs. B}} = \frac{1}{1 + 10^{(\text{Elo}_B - \text{Elo}_A) / 400}}
\end{equation}
\end{boxedformula}

\paragraph{Inter-Rater Agreement} The consistency of rankings among the human evaluators was measured using Kendall's W (Coefficient of Concordance)\footnote{Kendall, M. G., \& Babington Smith, B. (1939). The problem of m rankings. \emph{Annals of Mathematical Statistics, 10}(3), 275-287.}. Kendall's W assesses the level of agreement among several raters when ranking a set of items. It is calculated as follows:
\begin{boxedformula}
    \begin{equation}
        W = \frac{12S}{k^2(n^3 - n)} \quad \text{where} \quad S = \sum_{i=1}^{n} (R_i - \bar{R})^2
    \end{equation}
\end{boxedformula}
Here, $n$ is the number of items being ranked (the 6 methods), $k$ is the number of raters (evaluators), $R_i$ is the sum of the ranks assigned to method $i$ across all raters, and $\bar{R}$ is the mean of the $R_i$ values. The resulting score for our evaluation was \textbf{0.774}, indicating a strong level of agreement.

\paragraph{Top-Rank Frequency} In addition to Elo scores, we analyzed how frequently each method was ranked first by the human evaluators. The results, shown in Table \ref{tab:top_rank}, demonstrate a clear preference for the \methodname{} models.

\begin{table}[h!]
\centering
\caption{Top-Rank (\#1) Frequency in Human Evaluations.}
\label{tab:top_rank}
\begin{tabular}{@{}cc@{}}
\toprule
\textbf{Method} & \textbf{Top-Rank Percentage} \\
\midrule
\methodname{} (Gemini) & 46.54\% \\
\methodname{} (Gemma)  & 36.05\% \\
GPT-4o (ZS)          & 5.90\%  \\
LIVE                 & 4.87\%  \\
Gemini (ZS)          & 4.59\%  \\
Gemma (ZS)           & 2.06\%  \\
\bottomrule
\end{tabular}
\end{table}

\section{Additional Results and Analyses}
\subsection{Additional Qualitative Examples}
\begin{figure*}[h!]
    \centering
    \includegraphics[width=1\linewidth]{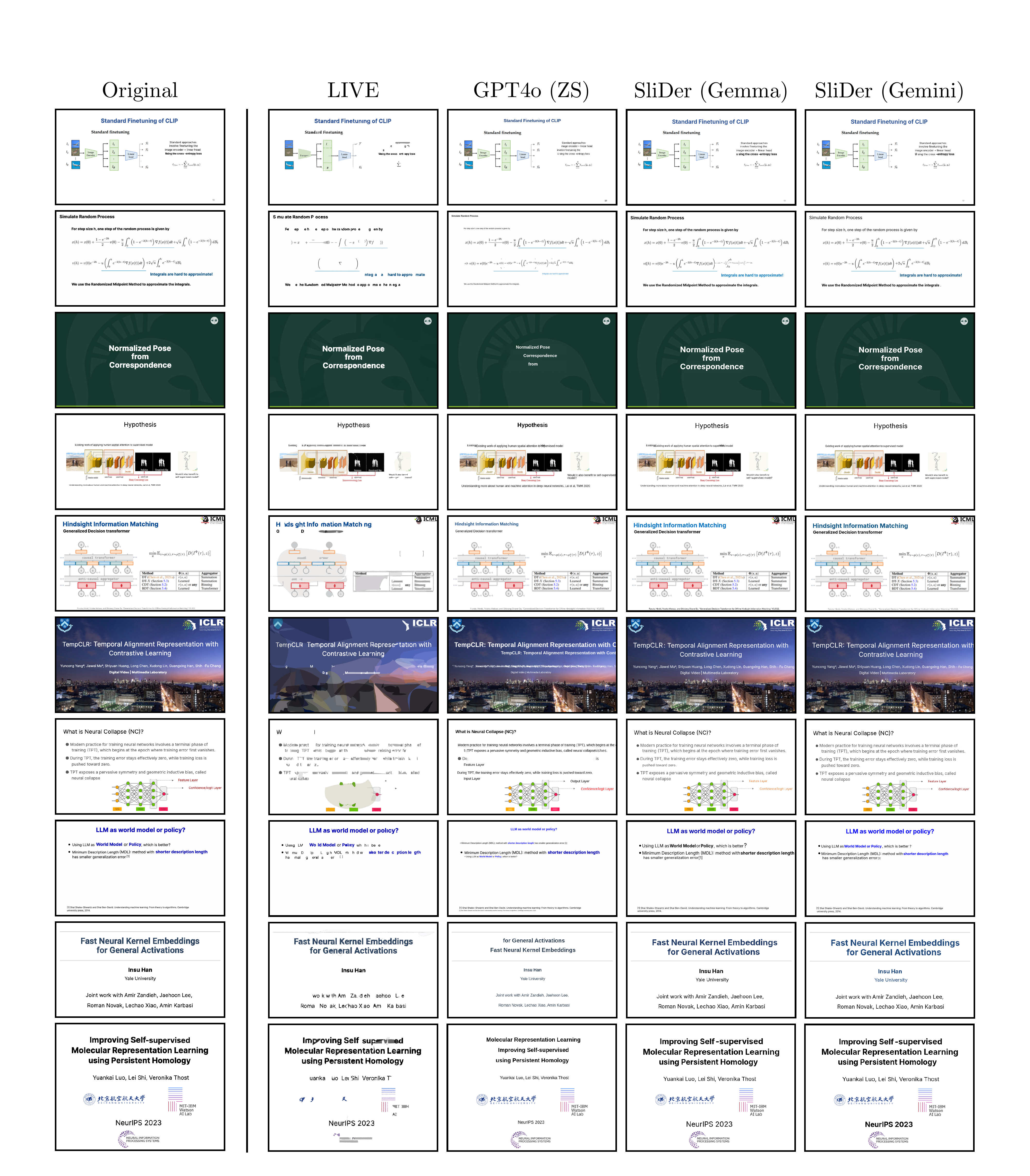}
    \caption{Additional examples of derendered slide images. Each row contains a separate sample, showing the original raster slide and the reconstructions produced by different methods. For \methodname{}, we show the YOLO-guided versions with one step of iterative refinement. ``ZS" refers to zero-shot methods.}
    \label{fig:appendix_examples}
\end{figure*}

Figure \ref{fig:appendix_examples} contains additional derendering examples similar to Figure 3 in the main paper, showcasing the performance of \methodname{} on a wider variety of slide layouts to demonstrate robustness.

\subsection{Ablation Study for Gemma-based \methodname{}}
Ablation results for the effect of YOLO bounding box priors and iterative refinement on Gemma-based \methodname{} are presented in Table \ref{tab:ablation_gemma}. The results mirror the findings for Gemini, confirming the significant benefits of using both YOLO-based spatial priors and iterative refinement (IR).

\begin{table}[h!]
\centering
\caption{Ablation results for Gemma-based \methodname{}, analyzing the impact of YOLO priors and iterative refinement (IR). \textbf{Bold} values correspond to the best results for each metric.}
\label{tab:ablation_gemma}
\resizebox{0.8\textwidth}{!}{%
\begin{tabular}{@{}c cc c c ccc@{}}
\toprule
\multirow{2}{*}{\textbf{Model}} & \multirow{2}{*}{\textbf{YOLO}} & \multirow{2}{*}{\textbf{IR}} & \multirow{2}{*}{\textbf{mIoU (\%) ↑}} & \multirow{2}{*}{\textbf{OCR Acc. (\%) ↑}} & \multicolumn{3}{c}{\textbf{Visual metrics}} \\
\cmidrule(l){6-8} 
& & & & & \textbf{MSE ↓} & \textbf{LPIPS ↓} & \textbf{CLIP Sim. ↑} \\
\midrule
\multirow{4}{*}{\shortstack{\textbf{\methodname{}}\\(Gemma)}} 
 & \textcolor{red!80!black}{\ding{55}} & \textcolor{red!80!black}{\ding{55}} & 69.64 & 91.83 & 15.49 & 0.105 & 0.916 \\
 & \textcolor{green!70!black}{\ding{51}} & \textcolor{red!80!black}{\ding{55}} & 89.04 & 92.90 & \textbf{13.34} & 0.076 & 0.949 \\
 & \textcolor{red!80!black}{\ding{55}} & \textcolor{green!70!black}{\ding{51}} & 69.22 & 91.85 & 15.54 & 0.105 & 0.917 \\
 & \textcolor{green!70!black}{\ding{51}} & \textcolor{green!70!black}{\ding{51}} & \textbf{89.36} & \textbf{93.53} & 13.38 & \textbf{0.075} & \textbf{0.950} \\ 
\bottomrule
\end{tabular}
}
\end{table}

\subsection{Demonstrating SVG Editability}
\label{sec:svg_editability}

A central motivation for \methodname{} is that its outputs are not custom or proprietary vector formats, but standard SVG files. This means that, once a slide has been derendered, users can directly modify it with off-the-shelf tools: for instance, desktop vector editors, browser-based SVG editors, or even plain text editors. Modern browsers can also render local SVGs, making it easy to iterate by editing the SVG text and reloading the page.

\begin{figure*}[h!]
    \centering

    \begin{minipage}[t]{0.48\textwidth}
        \textbf{(a) Original SVG snippet}\\[0.25em]
\begin{lstlisting}[
    language=XML,
    emph={font-family,font-size,font-weight,letter-spacing,color,text-align,Pretraining},
    emphstyle=\bfseries
]
...
  <foreignObject x="7.6%
                 width="20.6%
                 overflow="visible">
    <div xmlns="http://www.w3.org/1999/xhtml"
         style="font-family: Inter; font-size: 38px;
                font-weight: normal; letter-spacing: 0.0em;
                color: #000000; text-align: left;">
      <div>Pretraining</div>
    </div>
  </foreignObject>
...
\end{lstlisting}
    \end{minipage}\hfill
    \begin{minipage}[t]{0.48\textwidth}
        \centering
        \textbf{(b) Original raster}\\[0.25em]
        \includegraphics[width=\linewidth]{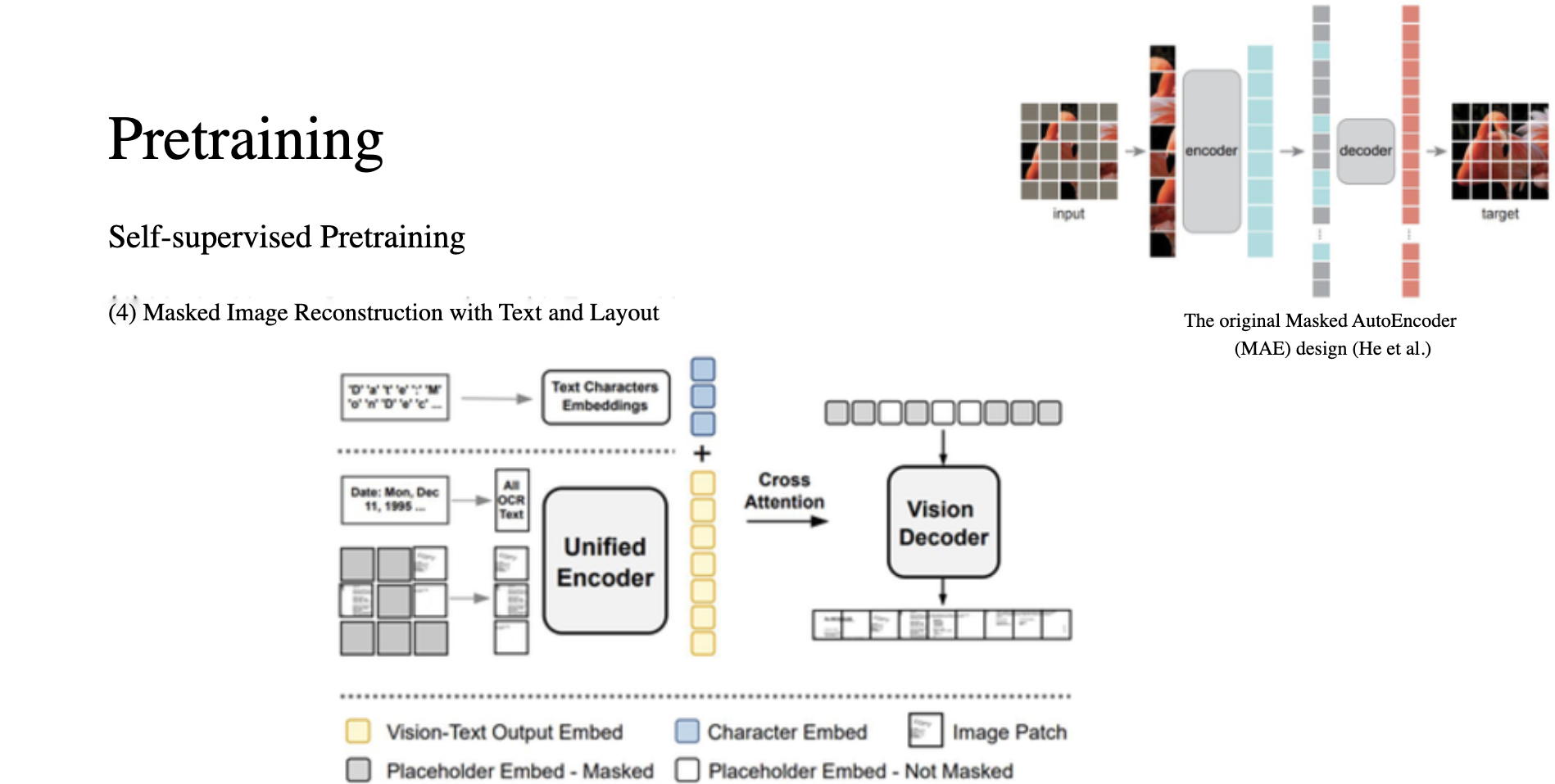}
    \end{minipage}

    \vspace{1em}

    \begin{minipage}[t]{0.48\textwidth}
        \textbf{(c) Modified SVG snippet}\\[0.25em]
\begin{lstlisting}[
    language=XML,
    emph={font-family,font-size,font-weight,letter-spacing,color,text-align,Modified,text},
    emphstyle=\bfseries
]
...
  <foreignObject x="20%
                 width="20.6%
                 overflow="visible">
    <div xmlns="http://www.w3.org/1999/xhtml"
         style="font-family: Inter; font-size: 48px;
                font-weight: bold; letter-spacing: 0.0em;
                color: blue; text-align: left;">
      <div>Modified text</div>
    </div>
  </foreignObject>
...
\end{lstlisting}
    \end{minipage}\hfill
    \begin{minipage}[t]{0.48\textwidth}
        \centering
        \textbf{(d) Modified raster}\\[0.25em]
        \includegraphics[width=\linewidth]{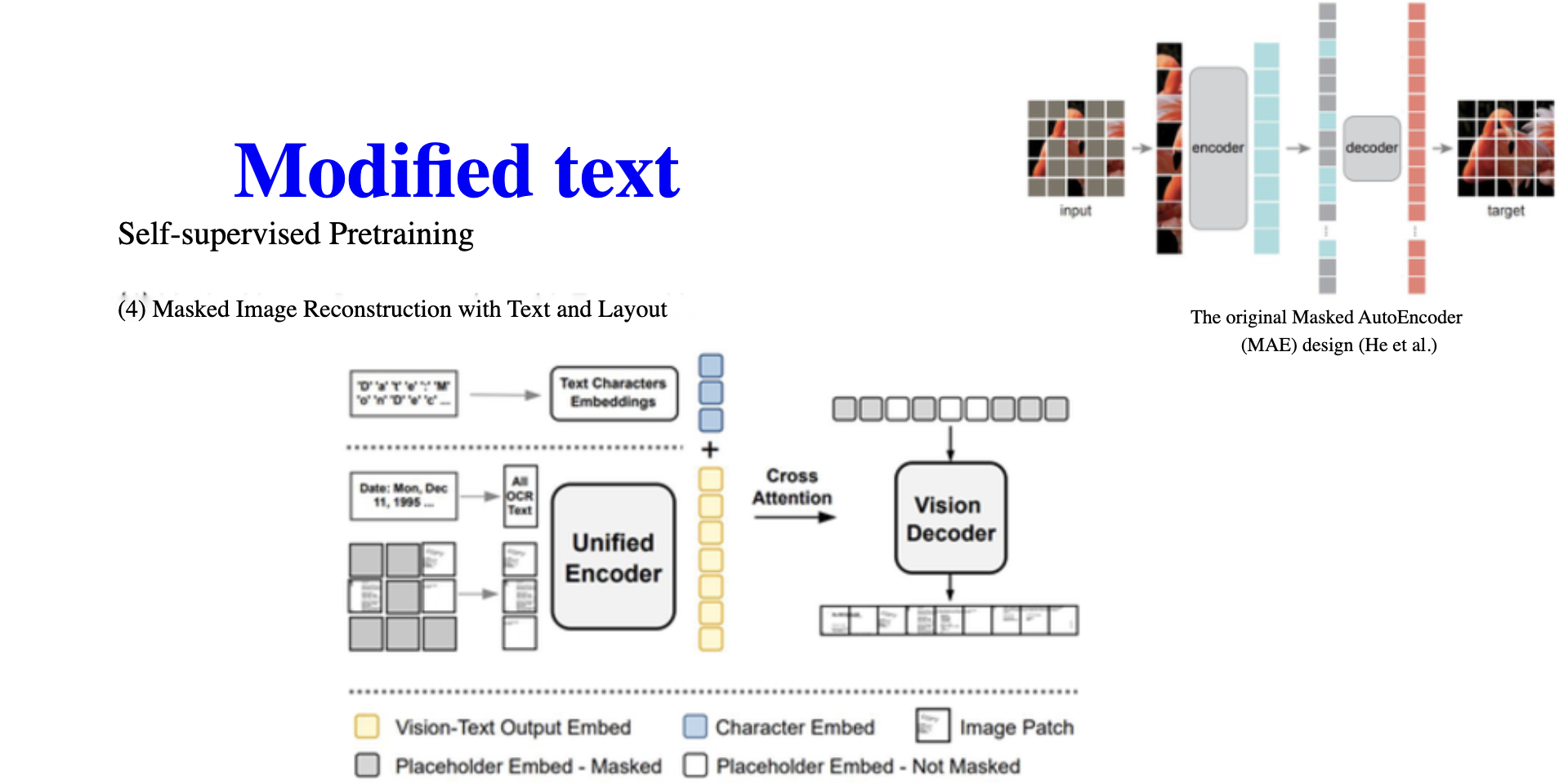}
    \end{minipage}

    \caption{
    Example of editing a \methodname{} output by directly modifying the SVG.
    The original snippet (a) renders to the raster in (b).
    Changing only a few attributes in the same \texttt{foreignObject} block
    (position, font size, font weight, color, and text content) yields the
    modified SVG in (c), which re-renders to (d).
    This illustrates that the outputs of our method are standard, easily
    editable SVGs that can be adjusted with simple
    text-based edits.
    }
    \label{fig:svg_editability}
\end{figure*}

Figure~\ref{fig:svg_editability} illustrates a simple example. The top row shows a small portion of an original SVG (left) and its raster rendering (right). The bottom row shows the result of editing only a few attributes in the same SVG snippet: we change the text position, font size, weight, color, and content. No special tools are required to perform these edits, as they correspond to directly modifying a single \texttt{foreignObject} block in the SVG text.

In practice, users can perform substantially richer edits than the example above: moving or deleting assets, swapping figures, or changing typography. Because all derendered slides are valid SVG documents, these edits can be carried out in widely available tools (e.g., vector graphics editors or web-based SVG editors) or directly in text editors and browsers, without any additional processing or post-hoc conversion.

\subsection{Framework Limitations and Future Work}
Our work presents a significant step towards semantic document derendering, but there are several limitations that offer avenues for future research:
\begin{itemize}
    \item \textbf{Supported SVG Elements:} The current implementation of \methodname{} exclusively handles text and image elements. Future work could involve expanding the \datasetname{} dataset to include primitive vector shapes (e.g., rectangles, paths), enabling the VLM to produce richer vector graphics.
    \item \textbf{Depth of Iterative Refinement:} Due to the high computational cost of generating training data for refinement, our models are trained with only a \textit{single} step. The framework itself supports multi-step refinement, which could further improve derendering quality.
    \item \textbf{Scope of Document Complexity:} This work focuses on slide documents. The same framework could be extended to derender more complex documents like posters, infographics, and scientific papers, provided a suitable dataset is available.
    \item \textbf{Training Depth:} All models were trained for only one epoch due to resource constraints. Performance could potentially improve with further training.
\end{itemize}

\end{document}